\documentclass{article}

\usepackage{arxiv}

\usepackage[utf8]{inputenc} 
\usepackage[T1]{fontenc}    
\usepackage{hyperref}       
\usepackage{url}            
\usepackage{booktabs}       
\usepackage{amsfonts}       
\usepackage{nicefrac}       
\usepackage{microtype}      
\usepackage{lipsum}
\usepackage{graphicx}
\usepackage{amsmath}
\graphicspath{ {./images/} }

\title{Uncovering Brain-Like Hierarchical Patterns in Vision-Language Models through fMRI-Based Neural Encoding}

\author{
 Yudan Ren \\
  School of Electronic and Information Engineering\\
  Northwest University\\
  Xi'an, Shaanxi 710127 \\
  \texttt{yudan.ren@nwu.edu.cn } \\
   \And
 Xinlong Wang \\
  School of Electronic and Information Engineering\\
  Northwest University\\
  Xi'an, Shaanxi 710127 \\
  \texttt{202332973@stumail.edu.cn} \\
  \And
 Kexin Wang \\
  School of Electronic and Information Engineering\\
  Northwest University\\
  Xi'an, Shaanxi 710127 \\
  \texttt{wangkx@stumail.nwu.edu.cn} \\
  \And
 Tian Xia \\
  School of Electronic and Information Engineering\\
  Northwest University\\
  Xi'an, Shaanxi 710127 \\
  \texttt{202323229@stumail.nwu.edu.cn} \\
  \And
 Zihan Ma \\
  School of Electronic and Information Engineering\\
  Northwest University\\
  Xi'an, Shaanxi 710127 \\
  \texttt{202233494@stumail.nwu.edu.cn} \\
  \And
 Zhaowei Li \\
  School of Electronic and Information Engineering\\
  Northwest University\\
  Xi'an, Shaanxi 710127 \\
  \texttt{202332984@stumail.nwu.edu.cn} \\
  \And
 Xiangrong Bi \\
  School of Electronic and Information Engineering\\
  Northwest University\\
  Xi'an, Shaanxi 710127 \\
  \texttt{bixiangrong@stumail.nwu.edu.cn} \\
  \And
 Xiao Li \\
  School of Electronic and Information Engineering\\
  Northwest University\\
  Xi'an, Shaanxi 710127 \\
  \texttt{lixiao@nwu.edu.cn} \\
  \And
 Xiaowei He \\
  School of Electronic and Information Engineering\\
  Northwest University\\
  Xi'an, Shaanxi 710127 \\
  \texttt{hexw@nwu.edu.cn} \\
}

\begin{document}

\maketitle

\begin{abstract}
While brain-inspired artificial intelligence (AI) has demonstrated promising results, current understanding of the parallels between artificial neural networks (ANNs) and human brain processing remains limited: (1) unimodal ANN studies fail to capture the brain's inherent multimodal processing capabilities, and (2) multimodal ANN research primarily focuses on high-level model outputs, neglecting the crucial role of individual neurons. To address these limitations, we propose a novel neuron-level analysis framework that investigates the multimodal information processing mechanisms in vision-language models (VLMs) through the lens of human brain activity, with comparisons to unimodal baselines (ViT/RoBERTa). Our approach uniquely combines fine-grained artificial neuron (AN) analysis with fMRI-based voxel encoding to examine two architecturally distinct VLMs: CLIP and METER. Our analysis reveals four key findings: (1) ANs successfully predict biological neuron (BN) activity across multiple functional networks (including language, vision, attention, and default mode), demonstrating shared representational mechanisms; (2) Both ANs and BNs show functional redundancy via overlapping representations, mirroring the brain's fault-tolerant processing; (3) ANs exhibit polarity patterns that parallel those in BNs, with oppositely activated BNs showing mirrored activation trends across VLM layers, reflecting the complexity and bidirectional nature of neural information processing; (4) The architectures of CLIP and METER drive distinct patterns of BN activation: CLIP's independent branches show modality-specific specialization, whereas METER's cross-modal design yields unified activation across modalities, highlighting the architecture's influence on ANN brain-like properties. These results provide compelling evidence for brain-like hierarchical processing in VLMs at the neuronal level, offering novel insights for developing more effective brain-inspired AI architectures.
\end{abstract}

\section{Introduction}
The human brain's remarkable efficiency inspires efforts to build AI with human-like decision-making and problem-solving abilities. Here, cognitive science provides essential theoretical guidance. For example, studying how the brain learns efficiently offers key insights for developing more powerful AI systems \cite{ohki2023efficient, wang2024brain}.

In recent years, ANNs represented by deep learning have made breakthroughs in the fields of computer vision and natural language processing \cite{radford2019language, Devlin2019BERTPO, raffel2020exploring, brown2020language, zhao2023brains, achiam2023gpt}. Moreover, as ANNs improve, their task processing efficiency continues to increase, with some models achieving human-level or even superior performance \cite{lake2023human}. These remarkable advancements have thus sparked growing interest in the similarities between ANNs and brain information processing. In fact, studies of multilayer perceptrons (MLPs), convolutional neural networks (CNNs), long short-term memory (LSTMs), and Transformers demonstrate that models with enhanced representational capabilities typically exhibit internal representations closer to neural responses. This drives exploration of hierarchical stimulus representations in ANNs versus human brains, where neural encoding models help elucidate natural stimulus-neural response relationships \cite{yamins2016using, le2025neural}.

Despite these promising developments, unimodal intelligent networks lack the brain's innate ability to integrate across modalities, while emerging multimodal architectures (e.g., VLMs) bridge this gap by modeling biological integration mechanisms. Although initial studies have examined the relationship between VLMs and brain activity, several critical issues remain unresolved. First, many of these studies rely on unimodal experimental stimuli (static images, text alone), which may not adequately reflect naturalistic multimodal perception \cite{wang2023better, tang2024brain}. Second, comparisons between VLMs and brain responses often lack rigorous unimodal baselines, making it unclear whether observed correspondences stem from genuine multimodal processing or the combined effects of strong unimodal components.

To address these challenges, we develop a fine-grained, neuronal-level encoding framework to investigate the alignment between VLM and brain mechanisms under multimodal perception, compared against unimodal benchmarks. Our approach begins by extracting fine-grained ANs from the multi-head attention modules of VLMs and unimodal baselines, followed by constructing their temporal responses. We subsequently apply sparse regression techniques to derive characteristic temporal activation patterns from these ANs, which serve as regressors in neural encoding models to predict brain activity (BN). Building upon this methodology, our neuronal-level analyses yield the following key findings. First, ANs in VLMs effectively predict BNs across diverse functional networks, including those for language, vision, attention, and the default mode, indicating their shared specialized networks for stimulus processing. Notably, VLMs outperform unimodal baselines in predicting brain networks within the multimodal correlation cortex, confirming their closer alignment with human cross-modal integration mechanisms. Second, VLMs exhibit functional redundancy and overlapping neural representations, mirroring the brain’s fault-tolerant processing architecture. Third, the polarity patterns of ANs (the distribution trends across VLM layers) exhibit aligned or opposing trends compared to those in BNs (the positive and negative  brain activation), underscoring the hierarchical and bidirectional information flow in both systems. Finally, architectural differences between VLMs significantly shape their brain-like characteristics: CLIP’s independent-branch architecture fosters modality-specific specialization, while METER’s cross-modal fusion promotes unified representations.

\section{Related Work}
\subsection{Consistency between ANNs and Human Brain Neuron response}
Recent years have witnessed substantial progress in examining the alignment between ANNs and human brain activity. These investigations primarily focus on the correspondence between ANN responses and neural activations during specific cognitive tasks. In visual processing, for instance, CNN representations have shown strong correlations with neural activity in the human visual cortex \cite{xu2021limits}. Similarly, the hierarchical organization of DNNs has been found to reflect the spatiotemporal dynamics of brain activity during object recognition, suggesting that ANNs can effectively model the dynamic nature of neural responses to visual stimuli \cite{cichy2016comparison}.

In natural language processing, word embedding models have demonstrated the ability to predict brain neural responses to linguistic stimuli, highlighting their capacity to emulate mechanisms of human language processing \cite{huth2016natural}. More recently, large language models (LLMs) have exhibited strong correspondences between their internal representations and neural activity during semantic tasks \cite{liu2023coupling, sun2024brain}. These findings collectively validate the use of ANNs as models of unimodal brain functions. Critically, however, the human brain operates fundamentally through multimodal integration. Although multimodal models outperform unimodal ones in predicting brain responses to audiovisual stimuli, their alignment advantages over unimodal baselines, especially at the fine - grained neuronal level, are still unexplored. To fill this gap, our study employs a unified neuron-level framework comparing VLMs against unimodal baselines.

\subsection{Neuron Representation Mapping with Multimodal Transformer Models}
Recent developments in LLMs have revealed remarkable reasoning abilities across both linguistic and visual domains \cite{zhao2023survey}. However, traditional LLMs are limited to processing discrete text, essentially overlooking visual information. The emergence of multimodal transformer models, particularly VLMs, has catalyzed new opportunities in neuroscience research \cite{lu2022multimodal, tang2024brain, Wang2022Natural}. Unlike unimodal neural networks, VLMs are designed to jointly process visual and textual information, enabling integration mechanisms that may resemble those of the human brain. For instance, CLIP \cite{radford2021learning} employs contrastive learning for cross-modal alignment, while METER \cite{dou2022empirical} incorporates explicit fusion layers for joint processing. Understanding how such architectural differences relate to brain function is crucial. Previous studies \cite{bavaresco2024modelling, subramaniam2024revealing} have demonstrated that multimodal models outperform unimodal models \cite{oota2025multi} in predicting fMRI responses in audiovisual tasks. However, these studies predominantly rely on embeddings derived from the final feedforward layers of target models. Although information-rich, such high-level embeddings may obscure critical fine-grained dynamics: specifically, interactions among processing units (e.g., attention heads) and layer-specific contributions throughout the network hierarchy.

What sets our work apart is: (1) A fine-grained definition of artificial neurons based on attention mechanisms (inspired by \cite{liu2023coupling}) to probe sublayer dynamics, providing more granular insights than layer embeddings. While exploratory, this enables investigation of finer representational details. (2) A systematic comparison of multimodal VLMs (CLIP and METER) against unimodal baselines (ViT and RoBERTa) within this framework. This design critically tests whether VLMs exhibit brain-like multimodal integration properties beyond their unimodal components, aiming to identify neuron-level signatures distinguishing multimodal from unimodal processing.

\section{Method}
\subsection{Overview}
\begin{figure*}[ht]
    \centerline{\includegraphics[width=\textwidth]{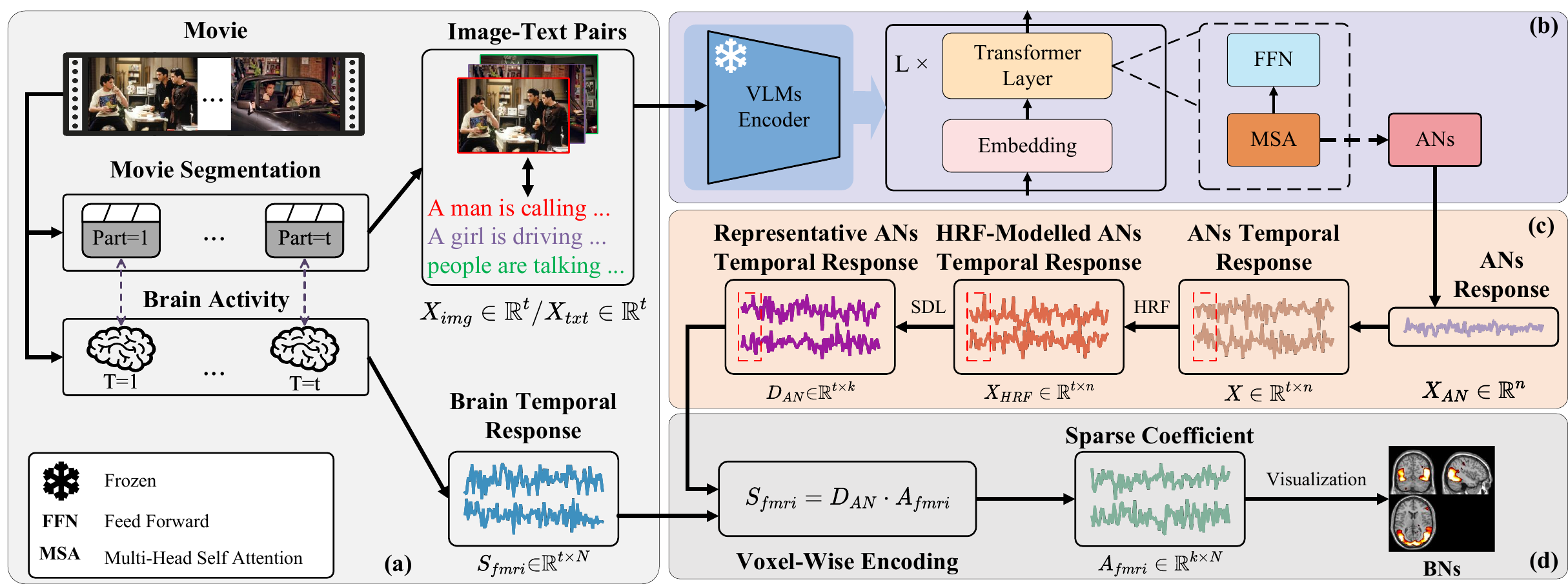}}
    \caption{Overview of the study. (a) Data Acquisition. (b) Construction of Fine-Grained ANs from Multi-Head Self-Attention Modules. (c) Extraction of Representative Activation Patterns of ANs via SDL. (d) Predicting BNs using Voxel-wise Encoding Model.}
    \label{fig1}
\end{figure*}

The overall framework is illustrated in Fig.~\ref{fig1}. Since the fMRI data are collected during the movie-watching session, we first preprocess the movie stimuli to ensure precise temporal alignment with the corresponding neural responses (Fig.~\ref{fig1}(a)). The details of alignment are provided in Section 4.1. We then construct fine-grained ANs from selected VLMs, as described in Section~3.2 (Fig.~\ref{fig1}(b)). To account for the hemodynamic delay inherent in fMRI recordings, we convolve the temporal response profiles of the ANs with a canonical hemodynamic response function (HRF), and use a sparse representation method (Sparse Dictionary Learning, SDL) to extract representative temporal activation patterns from these HRF-convolved responses (Fig.~\ref{fig1}(c)). These patterns are then used as regressors in a neural encoding model to predict voxel-level functional brain activity (Fig.~\ref{fig1}(d)). Finally, the predicted voxel responses are aggregated to construct functional brain networks (FBNs), which represent neural activity patterns of biological neurons (BNs) \cite{liu2023coupling}.

\subsection{ANs and Their Temporal Response in VLMs}
Defining fine-grained ANs that align with complex biological neural processes is a challenge. Inspired by recent work on the internal mechanisms of Transformer architectures \cite{liu2023coupling}, we construct ANs based on the multi-head attention module \cite{vaswani2017attention}. The standard attention score matrix $A$ in Transformer is computed as:
\begin{equation}
    A = \text{softmax}\left(\frac{QK^T}{\sqrt{d}}\right)
\end{equation}
where $Q \in \mathbb{R}^{n \times d}$ and $K \in \mathbb{R}^{n \times d}$ denote the query and key matrices for a given layer $l$ and attention head $h$, with $n$ representing the sequence length and $d$ the dimensionality of each head. The standard attention score $a_{jk}$ is computed via the dot product of the $j$-th query vector ($q_j \in \mathbb{R}^{d}$) and the $k$-th key vector ($k_k \in \mathbb{R}^{d}$). Simplifying by omitting the softmax and scaling factor, this dot product becomes:
\begin{equation}
    a'_{jk} = q_j \cdot k_k = \sum_{i=1}^{d} q_{ji} \cdot k_{ki}
\end{equation}
where $q_{ji}$ and $k_{ki}$ denote the $i$-th dimensional components of the vectors $q_j$ and $k_k$, respectively. Instead of summing the products across all dimensions $i$, we isolate the contribution of each dimension individually. For a specific dimension $i \in \{1, ..., d\}$, we compute an $n \times n$ matrix capturing the interactions between all query-key pairs through only that dimension. The $(j, k)$-th entry in this matrix corresponds to the scalar product $q_{ji}^{l,h} \cdot k_{ki}^{l,h}$. This interaction matrix for dimension $i$ is formulated as:
\begin{equation}
    \text{Matrix}_i^{l,h} = q_{:, i}^{l,h} (k_{:, i}^{l,h})^T
\end{equation}
where $q_{:, i}^{l,h} \in \mathbb{R}^n$ and $k_{:, i}^{l,h} \in \mathbb{R}^n$ are the vectors containing the $i$-th dimension across all query and key vectors, respectively.

\begin{figure}[ht]
    \begin{center}
        \centerline{\includegraphics[width=0.5\columnwidth]{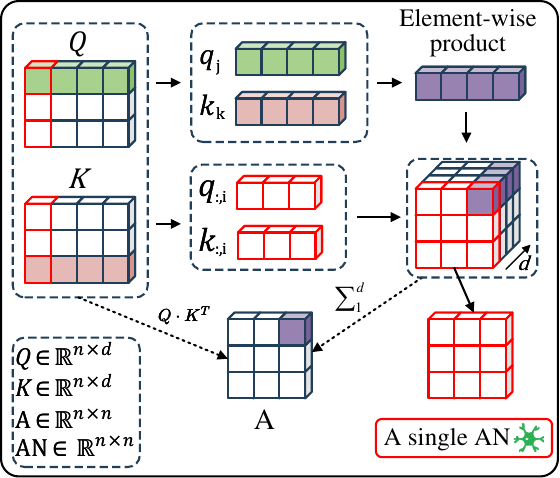}}
        \caption{Illustration of the AN construction process.}
        \label{fig2}
    \end{center}
\end{figure}

By performing this operation across all dimensions $i = 1, \dots, d$, we obtain $d$ distinct $n \times n$ matrices for each layer $l$ and head $h$. Each matrix represents the activity pattern of a single fine-grained AN, as conceptualized in Fig.~\ref{fig2}. The activation of this AN—associated with a specific $(l, h, i)$ tuple—is defined as the mean value of all elements in the corresponding matrix \cite{liu2023coupling}. This procedure results in a total of $N_L \times N_H \times d$ ANs per model, where $N_L$ is the number of attention layers and $N_H$ is the number of attention heads per layer. Applied to the architectures evaluated in this study, the total number of ANs is as follows:
\begin{itemize}
    \item \textbf{CLIP model:} 9216 ANs (vision encoder: 12 layers $\times$ 12 heads $\times$ 64 dimensions per head); 6144 ANs (text encoder: 12 layers $\times$ 8 heads $\times$ 64 dimensions per head)
    \item \textbf{METER model:} 4608 ANs per modality branch (6 layers $\times$ 12 heads $\times$ 64 dimensions per head)
    \item \textbf{ViT baseline:} 9216 ANs (12 layers $\times$ 12 heads $\times$ 64 dimensions per head)
    \item \textbf{RoBERTa baseline:} 9216 ANs (12 layers $\times$ 12 heads $\times$ 64 dimensions per head)
\end{itemize}

This consistent definition allows us to compare the internal dynamics derived from attention mechanisms across different model types, enabling the identification of potentially unique representational patterns associated with multimodal VLMs.

\subsection{Representative Temporal Response Patterns of ANs}
Having defined numerous fine-grained ANs, we now examine their temporal dynamics. Specifically, we analyze their activation sequences over $t$ time steps. However, directly analyzing the temporal activation patterns of such a large number of ANs individually poses significant computational challenges. Furthermore, direct observation of raw population activity may obscure meaningful underlying trends due to noise, redundancy, and potential autocorrelation effects \cite{sun2024brain}.

Motivated by neuroscience findings that suggest population-level neural representations often rely on a sparse set of core activity patterns \cite{daubechies2009independent, lv2015sparse}, we adopt a sparse coding approach to summarize the temporal behavior of the AN population. Specifically, we employ sparse dictionary learning \cite{mairal2009online, lv2015sparse} to identify a compact set of representative temporal response patterns from the full activity matrix. Let $X \in \mathbb{R}^{t \times n}$ denote the matrix where each column $X_{:,j}$ represents the temporal response (activation sequence over $t$ steps) of the $j$-th AN. Our goal is to learn a dictionary matrix $D_{AN} \in \mathbb{R}^{t \times k}$, whose $k$ columns serve as basis temporal patterns, and a sparse coefficient matrix $A_{AN} \in \mathbb{R}^{k \times n}$ indicating the contribution of each pattern to each AN. The optimization objective is given by:
\begin{equation}
    \min_{D_{AN}, A_{AN}} \|X - D_{AN}A_{AN}\|_{F}^2 + \lambda_{AN}\|A_{AN}\|_{1}
\end{equation}
where $\|\cdot\|_{F}$ denotes the Frobenius norm, $\|\cdot\|_{1}$ denotes the $L_1$ norm, and $\lambda_{AN}$ is a regularization parameter. Since $k \ll n$, the learned dictionary $D_{AN}$ effectively compresses the collective AN activity into a set of core temporal dynamics, reducing redundancy while retaining interpretability.

\subsection{Voxel-wise Encoding of Brain Activity}
To relate the temporal dynamics of ANs to BNs, we adopt a voxel-wise encoding model. This model uses the representative temporal patterns $D_{AN}$, obtained separately from each model type (VLMs and unimodal baselines), as regressors to predict voxel-level brain activity recorded by fMRI. This allows us to assess how well the neuronal patterns captured from each model type explain brain activity. Specifically, we employ a sparse coding framework, mathematically equivalent to LASSO regression, where the dictionary $D_{AN}$ (containing the AN patterns) is fixed. Let $S_{fmri} \in \mathbb{R}^{t \times N}$ denote the observed fMRI signal over $t$ time points and $N$ voxels. For each subject, we solve the following sparse regression problem:
\begin{equation} \label{eq:encoding}
    \min_{A_{fmri}} \|S_{fmri} - D_{AN}A_{fmri}\|_{F}^2 + \lambda_{fmri} \|A_{fmri}\|_{1}
\end{equation}
where $A_{fmri} \in \mathbb{R}^{k \times N}$ contains the learned coefficients for reconstructing voxel activity using the $k$ AN patterns. The regularization parameter $\lambda_{fmri}$ controls the sparsity of the coefficients, promoting interpretability by selecting only the most relevant temporal features for each voxel. Each entry $A_{fMRI,ij}$ reflects the contribution of the $i$-th temporal pattern to the $j$-th voxel's activity. This encoding model is trained separately for each subject. To evaluate group-level statistical significance, we perform a voxel-wise one-sample $t$-test across subjects on the learned coefficients $A_{fmri}$ for each temporal pattern in $D_{AN}$, assessing whether its contribution to voxel activity is significantly above chance. This analysis identifies brain regions that are significantly associated with specific representative AN-derived temporal patterns, thereby linking artificial model dynamics to biological neural populations.

\section{Experiments}
\subsection{Dataset and Preprocessing}
This study uses movie fMRI data from the publicly available Human Connectome Project (HCP) S1200 dataset, with details on experimental design and data acquisition provided in previous reports \cite{finn2021movie}. A total of 184 participants completed fMRI scans while watching short movie clips. These clips comprise both independently released videos (1.0–4.1 minutes, under Creative Commons licenses) and commercial Hollywood movie excerpts (3.7–4.3 minutes) \cite{cutting2012perceiving}. The fMRI data used in this study has undergone both spatial and temporal preprocessing, as provided by the HCP team. Additional preprocessing details can be found on the HCP website (https://db.humanconnectome.org/) and in related publications \cite{van2013wu, glasser2013minimal, finn2021movie, tian2021consistency}.

To ensure precise temporal alignment between neural responses and the movie stimuli, we perform the following steps:
\begin{enumerate}
    \item \textbf{Video segmentation:} The fMRI data comprise 4D time series, where each 3D brain volume (acquired per repetition time, TR) aligns with the movie segment viewed during that TR interval. This segmentation synchronizes the visual stimulus timeline with the acquired BOLD signal responses.
    \item \textbf{Frame selection:} Each segmented movie clip is processed to extract frames. Given the short duration of each segment (1 second) and minimal frame-to-frame variation, we select the first frame of each segment as the representative image—an approach widely adopted in neural encoding studies.
    \item \textbf{Text alignment:} Corresponding textual descriptions, provided by the Gallant Lab \cite{huth2012continuous}, are already temporally aligned with the fMRI recordings. 
\end{enumerate}

This temporal alignment synchronizes multimodal stimuli (image-text pairs) with fMRI volumes at fMRI's temporal resolution (TR). This ensures that multimodal inputs and resulting AN activations maintain temporal synchronization with neural recordings. This precise temporal correspondence enables us to investigate how ANNs process multimodal information in comparison to biological neural responses.

\subsection{Implementation Details}
To evaluate brain alignment capabilities in multimodal versus unimodal models, we compare four pre-trained models. The multimodal models include (1) CLIP (using the Hugging Face openai/clip-vit-base-patch32) \cite{radford2021learning} and (2) METER \cite{dou2022empirical}, which features parallel 6-layer vision and language Transformers. The unimodal baselines consist of (3) ViT, which shares the same architecture as CLIP’s vision encoder, and (4) RoBERTa-base \cite{liu2019roberta}, serving as a text-only comparator.

For preprocessing, input images are resized to $224 \times 224$ pixels and normalized according to the requirements of each model. Text inputs are tokenized using the model-specific tokenizers. These standardized inputs are then passed through their respective encoders to obtain internal representations. To ensure comparability, we apply a unified analysis pipeline across all four models for constructing ANs and mapping them to brain activity. We first extract query and key vectors from relevant layers as high-dimensional descriptors for AN construction. Then, we apply a consistent two-stage mapping framework to relate AN dynamics to BN activity (voxel-wise fMRI signals).

\begin{itemize}
    \item \textbf{Stage 1 (AN Feature Extraction):} We use Sparse Dictionary Learning (SDL) to extract $k=128$ representative temporal response patterns ($D_{AN}$) from the AN activity, employing a sparsity penalty $\lambda_{AN}=0.15$.
    \item \textbf{Stage 2 (Voxel Encoding):} We use these extracted AN patterns ($D_{AN}$) as features to predict voxel-wise fMRI activity ($X_{fmri}$) via LASSO regression, with a sparsity parameter $\lambda_{fmri}=0.2$.
\end{itemize}

This two-stage mapping, with hyperparameters fixed based on standard practices and preliminary checks, allows for a direct comparison of how well the neuronal patterns derived from each model (VLMs vs. unimodal) predict brain activity.

\section{Results}
\subsection{Applicability of the Sparse Representation in Artificial and Biological Neuron}
Neuroscience studies have demonstrated that neural processing in the brain exhibits sparse activation patterns, wherein only a small fraction of neurons are active at any given moment \cite{jaaskelainen2022sparse}. Drawing on this biological principle, we employ sparse coding techniques, specifically SDL, to transform the temporal activations of ANs into sparse representations. This approach aims to approximate biologically plausible encoding schemes and to improve the interpretability and efficacy of artificial models in predicting brain activity.

\begin{figure}[h]
    \begin{center}
    \centerline{\includegraphics[width=\columnwidth]{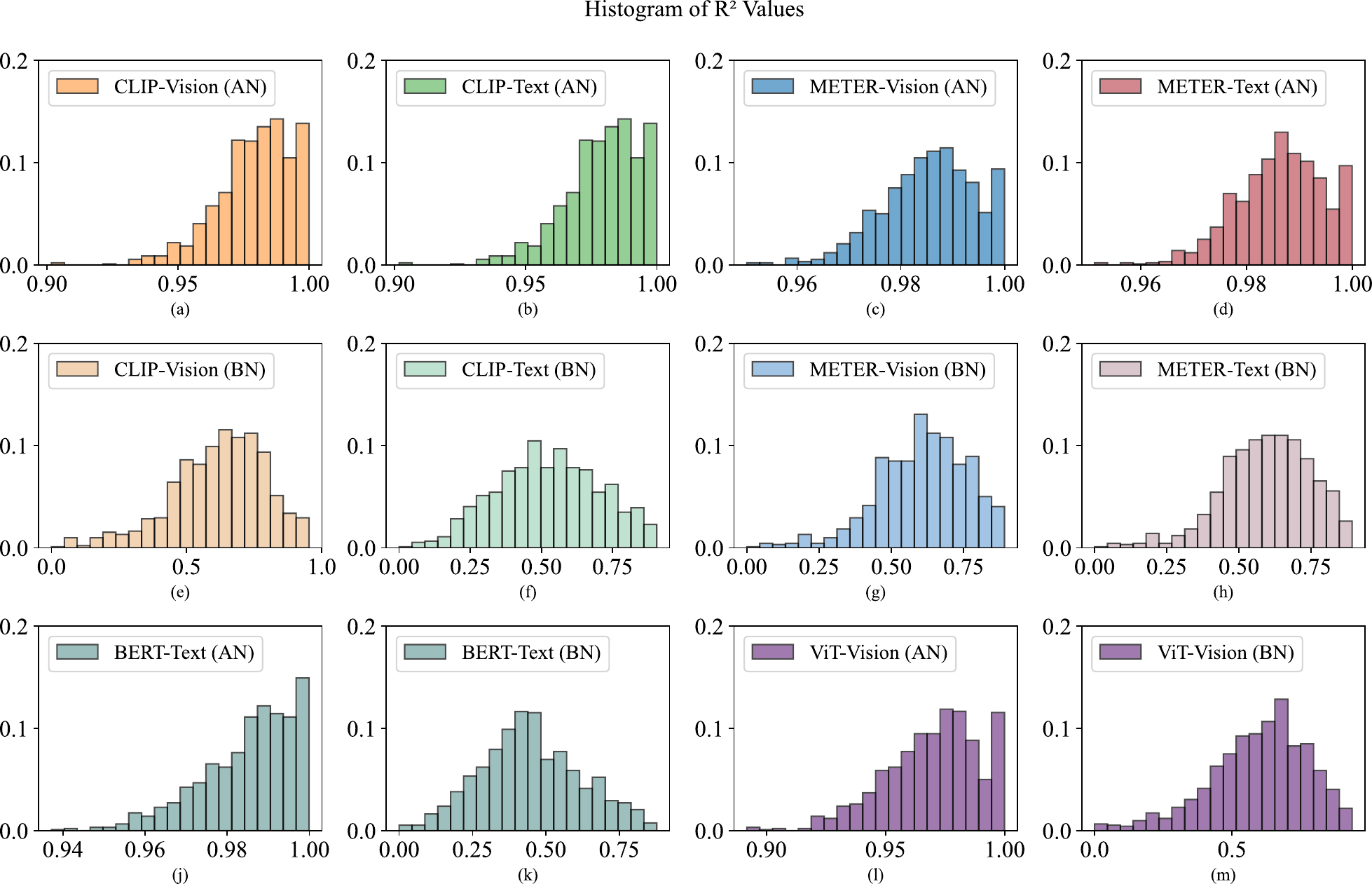}}
        \caption{The probability distribution of $R^{2}$ for different branches of models in ANs and BNs. (a)-(d), (j)-(k): $R^{2}$ distribution for ANs. (e)-(h), (i)-(m): $R^{2}$ distribution for BNs.}
        \label{fig3}
    \end{center}
\end{figure}

To assess the effectiveness of SDL in capturing AN dynamics and in predicting BN activity, we compute the coefficient of determination ($R^2$), which quantifies how well the learned sparse representations (derived from $D_{AN}$ and coefficients $A_{AN}$ or $A_{fmri}$) reconstruct the original AN temporal responses ($X$) or predict the voxel-wise fMRI signals ($S_{fmri}$), respectively. For each subject, we calculate the $R^2$ values for both AN and BN modeling and plot their distributions across subjects (Fig.~\ref{fig3}). All four models demonstrate high reconstruction fidelity for AN activity (Fig.~\ref{fig3}(a)–(d), (j)–(k)), with $R^2$ distributions strongly skewed toward higher values (median $>$ 0.95), indicating that sparse representations are adept at capturing the core informational constituents of ANNs, thus validating the utility of sparse coding. In contrast, the neural activation patterns of the brain are inherently more complex and heterogeneous compared to those of ANs, leading to the relatively lower $R^2$ values (median $\sim$0.6–0.7) (Fig.~\ref{fig3}(e)–(h), (i)–(m)). The distribution of BNs' $R^2$ values may reflect that certain BNs possess simpler activation patterns that fit well with dictionary $D_{AN}$, while others are more complex, implying that BNs' neural responses are more diverse.

To examine whether multimodal models improve brain alignment, we compare BN prediction $R^2$ values from VLMs against unimodal baselines. To validate this, two-sample $t$-tests on subject-level $R^2$ distributions reveal significant gains; notably, CLIP’s text branch (Fig.~\ref{fig3}(f)) outperforms RoBERTa (Fig.~\ref{fig3}(m)) ($p<0.05$), indicating VLMs encode richer representations aligned with multimodal neural dynamics.

Comparing CLIP and METER reveals different $R^2$ distributions: CLIP exhibits larger vision-text discrepancies in both AN reconstruction (Fig.~\ref{fig3}(a) vs. (b)) and BN prediction (Fig.~\ref{fig3}(e) vs. (f)), while METER shows consistent cross-branch $R^2$ distributions (Fig.~\ref{fig3}(c, d, g, h)). This divergence originates from their divergent alignment strategies: CLIP aligns global embeddings contrastively, potentially limiting local feature synergy, whereas METER employs cross-modal fusion layers that enhance localized interactions \cite{dou2022empirical}. During the multimodal information processing, the brain typically employs two complementary mechanisms: independent processing and cross-modal integration \cite{grossberg2000complementary}. CLIP and METER, to varying degrees, emulate these distinct neural processing strategies. CLIP's independent encoding resembles early-stage independent sensory processing in the brain, which facilitates the formation of high-resolution feature representations for each modality. In contrast, METER underscores the convergence of image and textual information, mirroring the synchronous modality processing observed in the brain's multimodal integration zones.

\subsection{Mapping between Sparse Dictionary and Brain Regions}
\begin{figure}[h]
\begin{center}
\centerline{\includegraphics[width=\columnwidth]{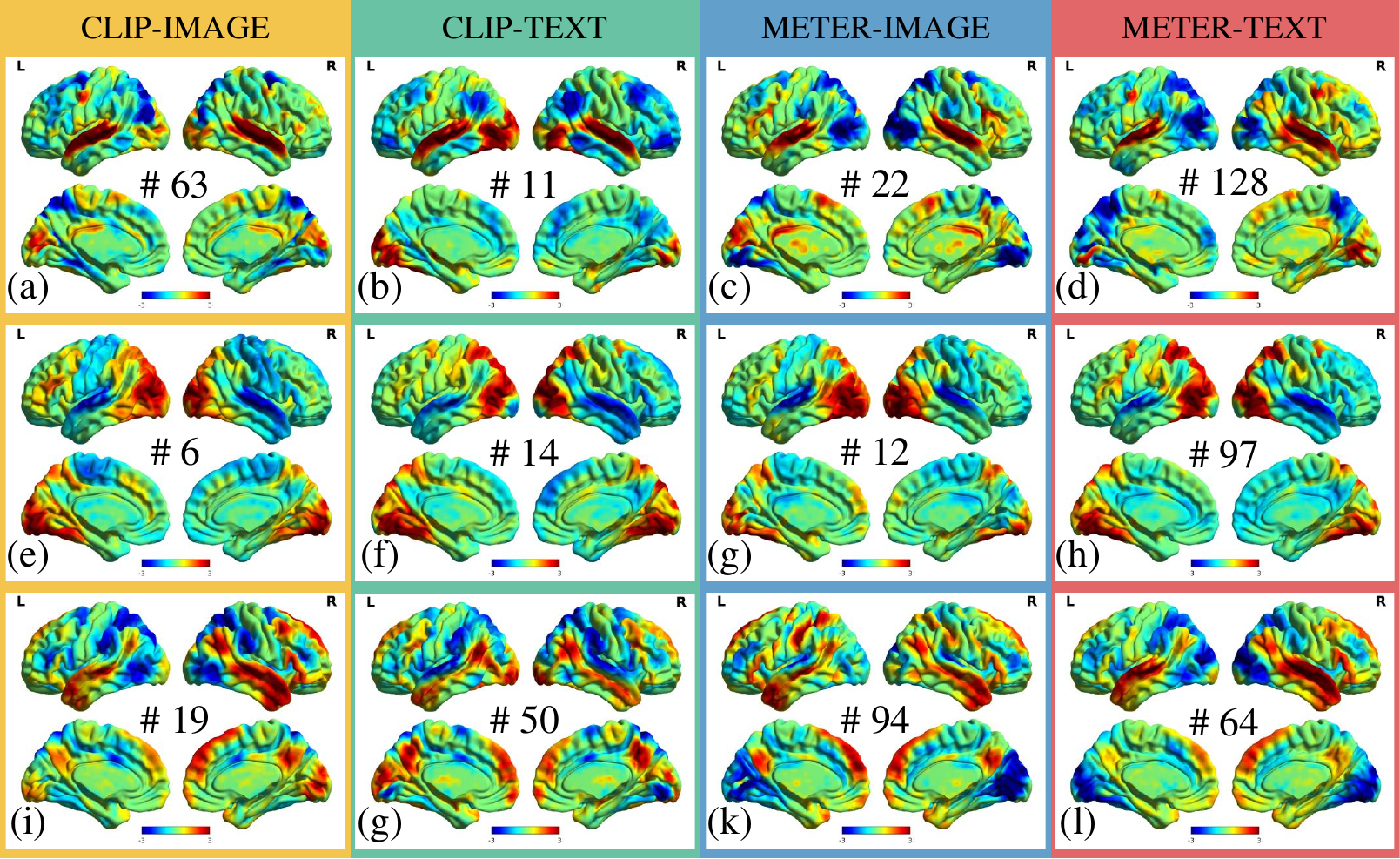}}
\caption{Brain activation maps corresponding to example dictionary atoms. Color intensity represents the t-statistic from the group-level one-sample t-test on the LASSO coefficients (\(A_{fmri}\) in Eq. \ref{eq:encoding}) for that atom across subjects, thresholded for significance (FDR corrected p < 0.05). Warm colors indicate positive coefficients (atom positively predicts activity); cool colors indicate negative coefficients. Maps are displayed on an inflated cortical surface. Examples shown activate regions including the auditory cortex (a-d), visual cortex (e-h), attention networks (e-h), default mode network (i-l), and language-related areas (i-l).}
\label{fig4}
\end{center}
\end{figure}

By projecting AN dynamics onto BNs, we observe significant overlap in BNs related to the text and visual branches of the VLM across multiple brain regions. These regions include the auditory cortex (Fig.~\ref{fig4}(a)-(d)), visual cortex (Fig.~\ref{fig4}(e)-(h)), attention network (Fig.~\ref{fig4}(e)-(h)), default mode network (Fig.~\ref{fig4}(i)-(l)), language-related regions (Fig.~\ref{fig4}(i)-(l)), as well as key multimodal integration areas like the anterior temporal lobe (ATL). These preliminary findings (as exemplified by the sample dictionary atom activation maps in Fig.~\ref{fig4}) suggest that these multimodal models have the potential to simulate the underlying neural mechanisms of cross-modal integration.

\begin{figure}[ht]
\begin{center}
\centerline{\includegraphics[width=0.65\columnwidth]{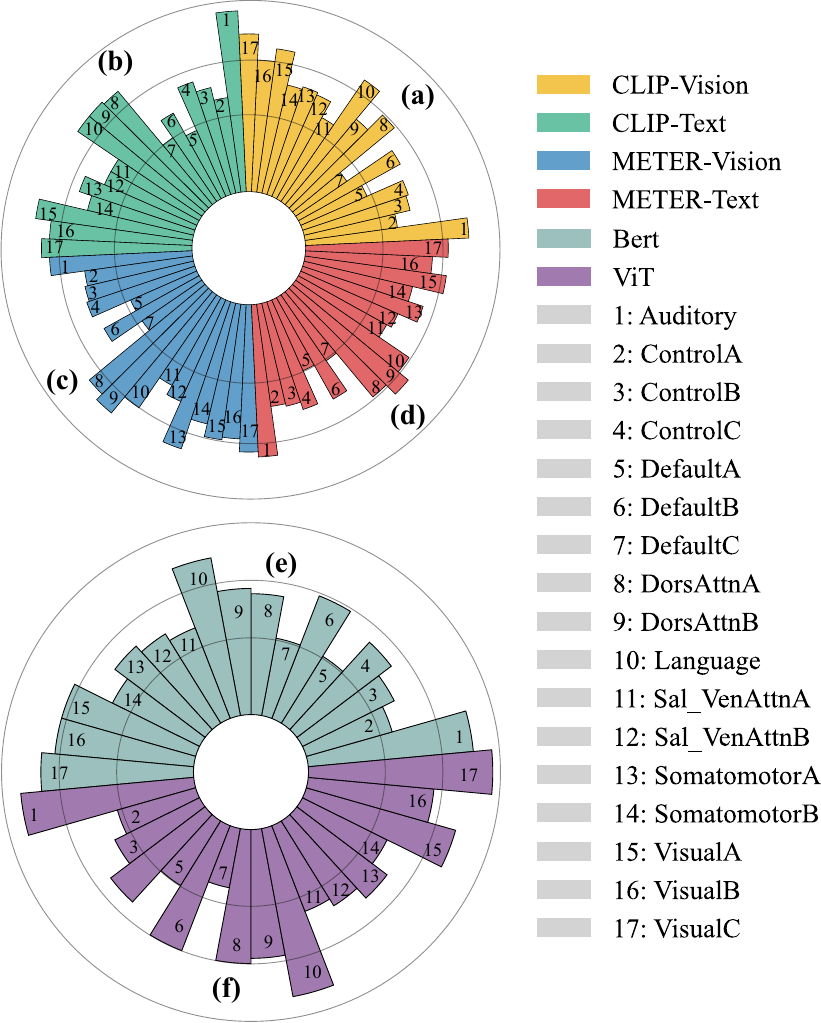}}
    \caption{Number of each functional parcellation in 128 BNs. 128 is the number of dictionary derived from ANs.}
    \label{fig5}
\end{center}
\end{figure}

Despite some overlap in brain regions, the text and vision branches exhibit distinct functional specialization, reflecting the different natures of the information processed. To quantify this, we compute Dice coefficients between specific brain regions within BNs and 17 predefined parcellations \cite{kong2021individual}, defining significant activation at $\text{Dice}>0.7$. For each model branch (CLIP-Visual, CLIP-Text, METER-Visual, METER-Text, RoBERTa-Text, ViT-Vision), we generated 128 BN maps from 128 dictionary atoms. We count the number of BNs significantly activating each functional parcellation per branch (up to 128). As shown in Fig.~\ref{fig5} (where different colors represent different model branches and bar height indicates the number of BNs activating the region), unimodal models demonstrate expected functional specialization: ViT atoms are primarily associated with the visual cortex (Fig.~\ref{fig5}(f)), while RoBERTa atoms show strong correlation with the language network (Fig.~\ref{fig5}(e)). CLIP's visual (Fig.~\ref{fig5}(a)) and text (Fig.~\ref{fig5}(b)) branches largely mirror this unimodal segregation pattern. The visual branch significantly activates visual areas, supporting that its visual encoding relies on known neural pathways \cite{kausel2024multimodal, wang2023better}; the text branch primarily activates language-related regions (such as the superior temporal gyrus, angular gyrus, Broca's area), consistent with existing research \cite{caucheteux2022brains}. This clear separation observed in CLIP simulates the characteristic functional specialization found in the human brain \cite{kriegeskorte2018cognitive}. In contrast, METER's visual (Fig.~\ref{fig5}(c)) and text (Fig.~\ref{fig5}(d)) branches appear to exhibit more similar activation patterns, showing comparable activation levels across visual, auditory, attention, and language understanding regions. This highly overlapping activation pattern might stem from METER's architectural design, which employs stronger cross-modal interaction mechanisms to integrate visual and language representations in a joint feature space \cite{caucheteux2022brains}. Such a mechanism promotes cross-modal integration by reducing reliance on modality-specific pathways, potentially reflecting the synergistic working characteristics of multimodal brain areas.

Overall, these results reveal the complexity of VLMs in simulating human brain information processing. On the one hand, VLMs (especially CLIP) can capture brain functional specialization related to specific sensory inputs (visual or language), similar to unimodal models (ViT, RoBERTa). On the other hand, VLMs also demonstrate the ability to simulate multimodal integration, as evidenced by the overlapping activation across modalities and the more integrated patterns displayed by METER, a capability that unimodal models cannot achieve alone. The observed overlap in cross-modally associated brain regions suggests that VLMs might spontaneously discover integration mechanisms similar to biological systems. This neurobiological correspondence depends on model architecture: CLIP's separated streams reproduce the functional specialization characteristic of unimodal sensory cortices, while METER's cross-modal architecture captures the integration dynamics found in heteromodal association areas. Together, these findings indicate that, compared to unimodal baselines, VLMs can better simulate key aspects of multimodal information processing in the human brain, and that the architecture of ANNs is a critical determinant for the emergence of brain-like multimodal integration computational principles.

\subsection{Feature Similarity and Redundancy Among Dictionary Atoms}
The brain achieves robust information processing through redundant neural structures. This biological principle of functional redundancy is considered a key feature of efficient and reliable brain function \cite{luppi2022synergistic} and enhances the flexibility of information processing.

\begin{figure}[ht]
    \begin{center}
    \centerline{\includegraphics[width=0.65\columnwidth]{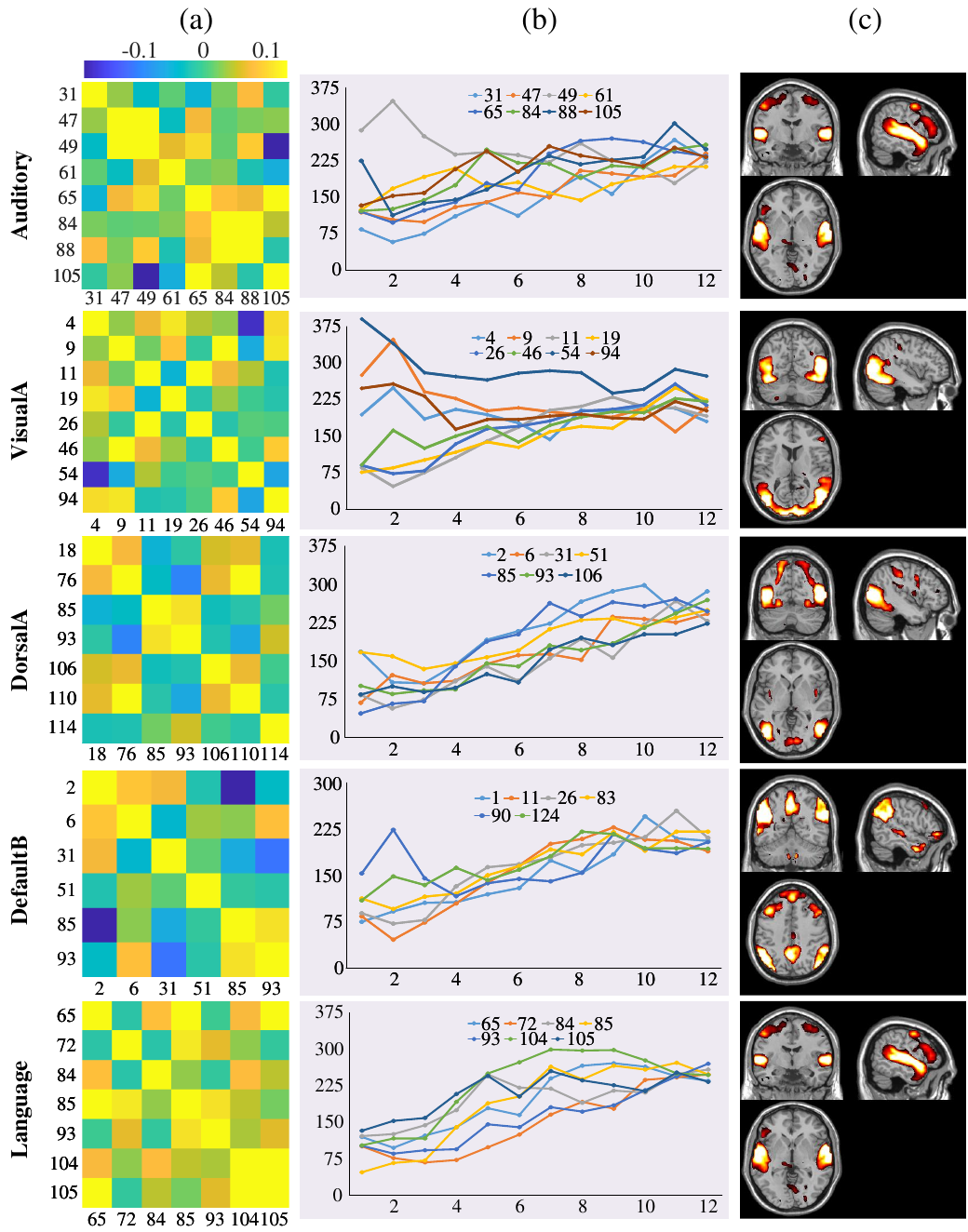}}
        \caption{Multiscale redundancy analysis of dictionary atoms. (a) Inter-atom correlation matrix of temporal activation patterns (rows of $D_{AN}$), where high correlations indicate shared feature dynamics. (b) Layer-wise distribution profiles of associated AN activations across VLM's transformer hierarchy. (c) Spatial overlap of BN activation maps within template-defined functional regions (e.g., auditory network masks).}
        \label{fig6}
    \end{center}
\end{figure}

To explore whether VLMs have similar redundancy mechanisms, we analyze the relationship between BNs and their corresponding ANs within the learned sparse dictionary in $D_{AN}$. Our initial observation reveals redundancy at the BN level: Following the activation criteria defined in Section 5.2, we evaluate which predefined functional brain regions are significantly activated by each BN. We observe that multiple BNs generated from different dictionary atom mappings frequently activate the same functionally partitioned regions. Fig.~\ref{fig6}(c) illustrates multiple BNs (corresponding to dictionary atoms indicated by IDs on the axes of Fig.~\ref{fig6}(a)) activating the same functional region (e.g., the auditory region shown in the "Auditory" row example). This overlap indicates that multiple BNs map to similar functional brain areas (ranging from visual networks to language networks), suggesting a similarity to the inherent redundancy of the brain. Having established redundancy at the BN level, we then investigate whether a similar redundancy exists at the level of ANs within the VLMs. We focus on the dictionary atoms in $D_{AN}$ associated with the previously identified overlapping BNs. Specifically, we first examine the correlation between the temporal activation patterns (rows of $D_{AN}$) of these atoms (Fig.~\ref{fig6}(a)) to assess the similarity of their representations. Additionally, using the matrix $A_{AN}$ which captures the relationship between each dictionary atom and ANs across all VLM layers, we analyze how these redundant atoms associate with ANs situated at different depths in the model's hierarchy. Our aim is to examine whether atoms mapping to the same functional brain region also exhibit similar layer-wise association patterns across the Transformer layers (Fig.~\ref{fig6}(b)). The observed inter-atom correlations (Fig.~\ref{fig6}(a)), coupled with similar AN activation distributions across VLM layers (Fig.~\ref{fig6}(b)), indicate that dictionary atoms predicting overlapping BN activity patterns also tend to correspond to ANs within the VLM that have similar feature representations and activation patterns. This convergence observed at the AN level demonstrates that redundancy in the VLM's internal representations mirrors principles observed in biological neural networks and contributes to robust multimodal information integration in ANNs.

\subsection{Polarity distribution of atomic activations and hierarchical properties of ANs}
\begin{figure}[h]
    \begin{center}
    \centerline{\includegraphics[width=\columnwidth]{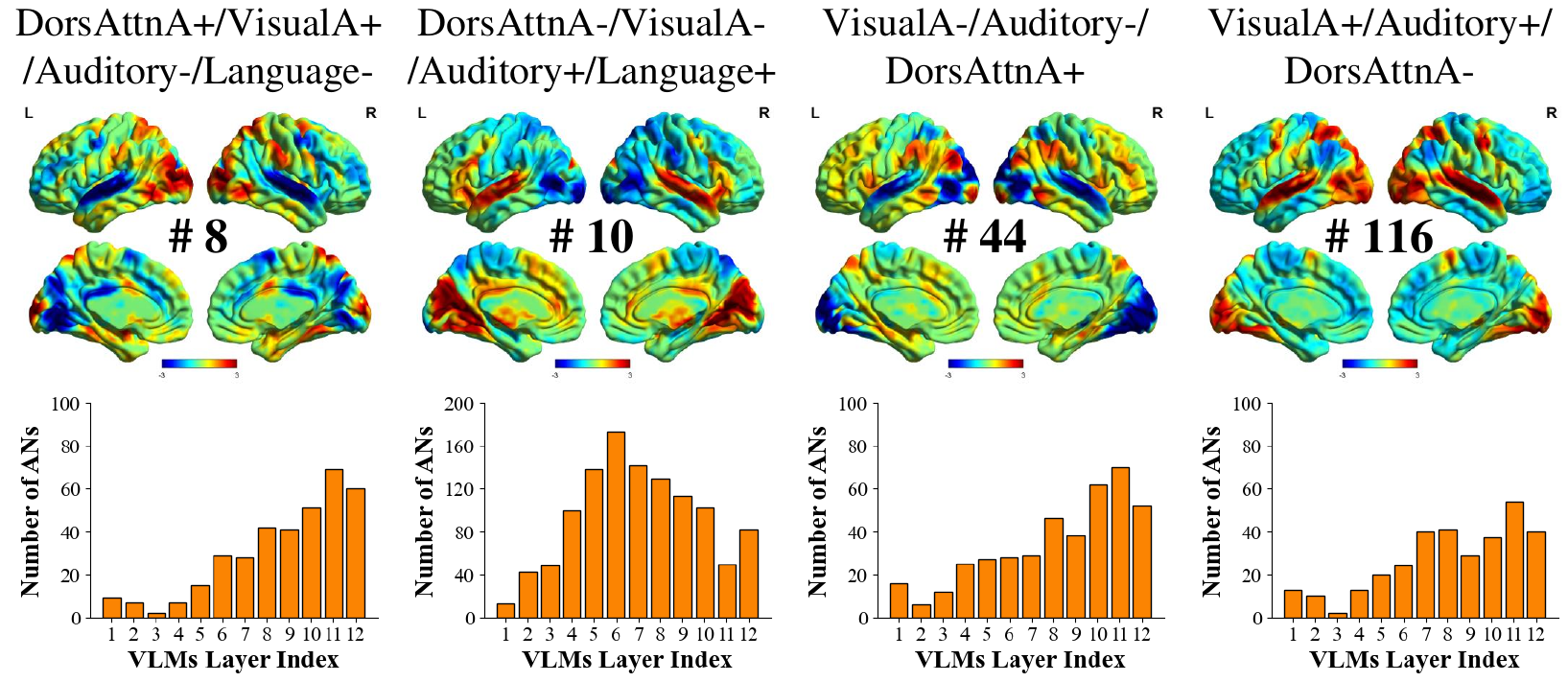}}
        \caption{Cross-modal analysis of BN polarity and AN layer distribution. (Upper) BN activation maps for atom pairs (8 vs. 10, 44 vs. 116) showing antagonistic polarities in core brain networks. (Lower) Corresponding AN activation profiles across VLM layers reveal layer-dependent polarity determinants. Layer numbering follows the standard VLM architecture.}
        \label{fig7}
    \end{center}
\end{figure}

Our previous studies demonstrate that the sparse dictionary $D_{AN}$ derived from ANs can reveal BNs with functional roles. However, the observed AN-BN relationships suggest bidirectional interactions rather than unidirectional mappings, motivating quantitative analysis of the polarity distributions to decode these complex dynamics (see polarity distributions in Fig.~\ref{fig7}).

Specifically, we analyze polarity-aligned activation patterns of ANs and their corresponding BNs. In the VLM's layers, atoms 8 and 10 produce BNs with antagonistic activations (excitatory vs. inhibitory) in cognitively relevant regions including DorsAttnA, VisualA, Auditory, and Language. Crucially, their associated ANs exhibit mirrored layer distributions - ANs for atom 8 peak in deeper layers, while atom 10's ANs dominate mid-anterior layers. This inverse correlation suggests layer-specific processing hierarchies in ANs directly modulate BN polarity. 

Conversely, atoms 44 and 116 generate BNs with opposing polarities despite their ANs sharing near-identical mid-to-late layer distributions. This dissociation implies multiple mechanisms govern polarity formation. Our systematic mapping of AN-BN polarity relationships provides compelling evidence for brain-inspired functional hierarchies in VLMs, demonstrating how sparse representations encode biologically plausible neural computations.

\section{Conclusion and Discussion}
This study introduces a fine-grained, neuron-level fMRI encoding framework to investigate the potential associations between ANs in VLMs and BNs in the human brain during the processing of complex, naturalistic multimodal stimuli. By systematically comparing VLMs with high-performance unimodal baseline models, we address the following question: whether and how VLMs emulate the organizational principles of cross-sensory information integration observed in the human brain.

The results demonstrate that while VLMs share fundamental brain-like properties with unimodal models, such as hierarchical representation structure, functional redundancy, and polarity correspondence between ANs and BNs. However, VLMs also exhibit distinct dynamic integration patterns when processing combined visual-linguistic inputs. Notably, VLM-derived features yield higher accuracy in predicting brain activity compared to unimodal baselines, suggesting that these models implement representational strategies compatible with the brain’s multimodal integration mechanisms.

Further architectural comparisons among VLMs reveal how design choices influence brain-like response patterns. Specifically, CLIP’s separated encoders tend to correlate with activity patterns in unimodal brain pathways (e.g., visual or language regions), resembling the brain’s functionally specialized processing pathways. In contrast, METER’s explicit cross-modal fusion architecture produces integrated activity patterns more closely aligned with heteromodal association cortices (e.g., the temporoparietal junction). These observations imply that distinct architectural strategies in VLMs capture different facets of the brain's multimodal processing system: separated encoders reflect functional segregation, while fusion mechanisms emulate cross-modal synergy.

This study encompasses several limitations. First, our attention-based definition of ANs, while offering fine-grained interpretability, requires further validation against alternative criteria (e.g., activation magnitude or connectivity strength) to ensure robustness. Second, the current analysis is limited to a small set of VLM architectures. Expanding to a broader array of models will be crucial for assessing the generalizability of our findings. Finally, translating these brain-inspired insights into actionable design principles for artificial systems will require interdisciplinary collaboration and focused experimental validation.

Despite these limitations, this study provides evidence at a fine-grained neuronal level for potential alignment between visual-language models and human multimodal processing mechanisms. We propose that the diverse architectures of VLMs can serve as a controllable computational platform for exploring the computational foundations of multimodal integration principles in biological systems. This perspective offers bidirectional implications: it inspires neuroscience-informed AI model design while simultaneously providing novel analytical tools for understanding the brain’s complex cognitive functions.

\bibliographystyle{unsrt}  





\bibliography{references}

\end{document}